%% file: main.tex
\documentclass[10pt,twocolumn,letterpaper]{article}

\usepackage[pagenumbers]{iccv} 

\definecolor{iccvblue}{rgb}{0.21,0.49,0.74}
\usepackage[pagebackref,breaklinks,colorlinks,allcolors=iccvblue]{hyperref}

\def\paperID{*****} 
\def\confName{ICCV}
\def\confYear{2025}

\title{Combining Transformers and CNNs for Efficient Object Detection in High-Resolution Satellite Imagery}

\author{Nicolas Drapier\\
L2TI Laboratory, Institut Galilée\\
Université Sorbonne Paris Nord\\
SAS Impact \\
{\tt\small nicolas.drapier@edu.univ-paris13.fr, nicolas.drapier@sas-impact.fr}
\and
Aladine Chetouani\\
L2TI Laboratory, Institut Galilée\\
Université Sorbonne Paris Nord\\
{\tt\small aladine.chetouani@univ-paris13.fr}
\and
Aurélien Chateigner\\
SAS Impact\\
{\tt\small aurelien.chateigner@sas-impact.fr}
}

\begin{document}
\maketitle
\input{sec/0_abstract}

\input{sec/1_2_intro_related_work}
\input{sec/3_methodology}

\input{sec/4_experiments}

\input{sec/5_conclusion}
{
    \small
    \bibliographystyle{ieeenat_fullname}
    \bibliography{main}
}

\end{document}

%% file: sec/0_abstract.tex
\begin{abstract}
We present GLOD, a transformer-first architecture for object detection in high-resolution satellite imagery. GLOD replaces CNN backbones with a Swin Transformer for end-to-end feature extraction, combined with novel UpConvMixer blocks for robust upsampling and Fusion Blocks for multi-scale feature integration. Our approach achieves 32.95\% on xView, outperforming SOTA methods by 11.46\%. Key innovations include asymmetric fusion with CBAM attention and a multi-path head design capturing objects across scales. The architecture is optimized for satellite imagery challenges, leveraging spatial priors while maintaining computational efficiency.
\end{abstract}

%% file: sec/1_2_intro_related_work.tex
\section{Introduction}
\label{sec:introduction}

\begin{figure}[t]
    \centering
    \includegraphics[width=\linewidth]{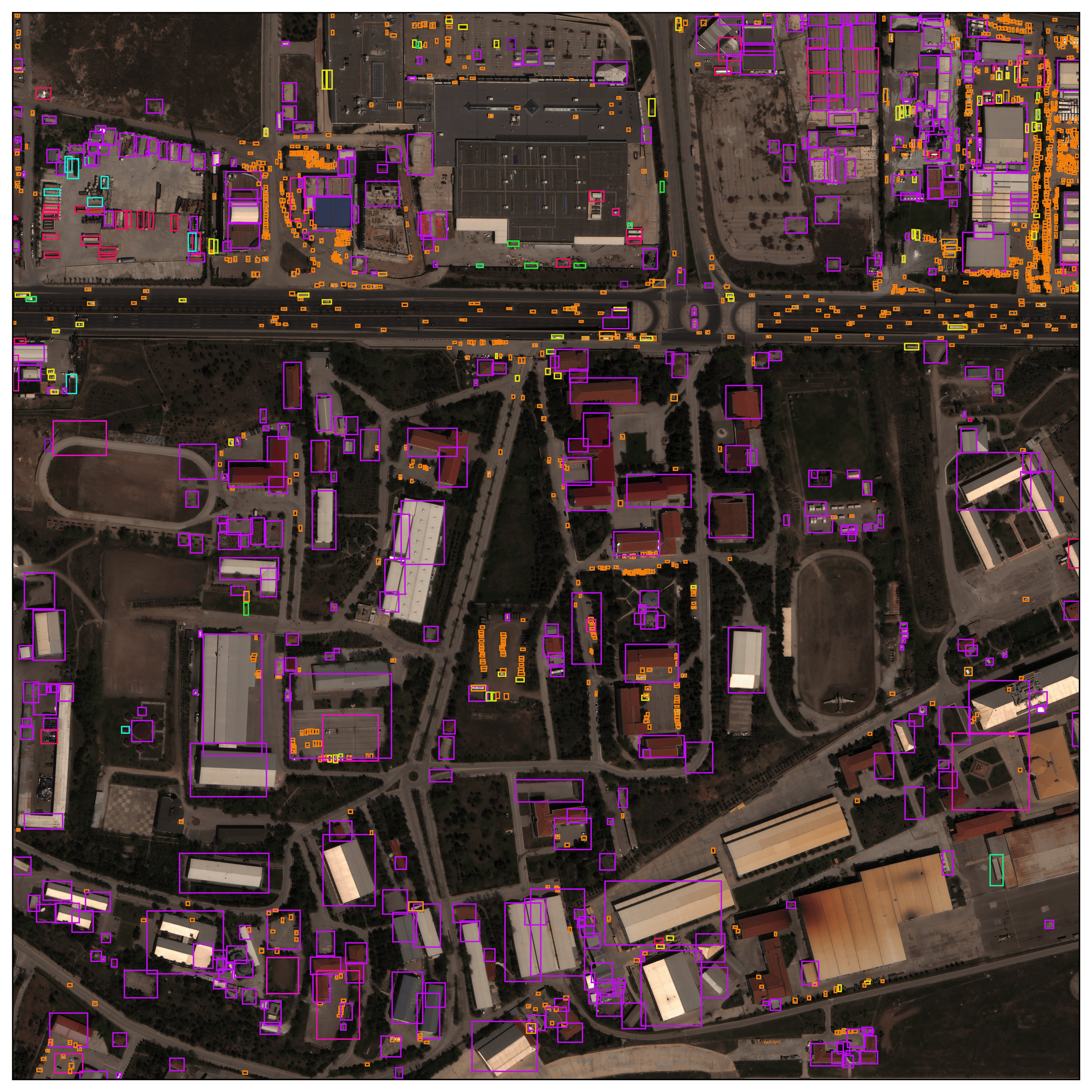}
    \caption{\textbf{High-resolution satellite image with dense object predictions from the xView dataset.} The scene highlights the challenges of small object detection in cluttered environments, including high object density, overlapping instances, and scale variation. Orange = \textit{Small cars}, Purple = \textit{Buildings}, Red = \textit{Container}, Green = \textit{Bus}, Cyan = \textit{Yacht}.}
    \label{fig:prediction_sample}
\end{figure}

\begin{figure*}[ht]
    \centering
    \includegraphics[width=0.85\textwidth]{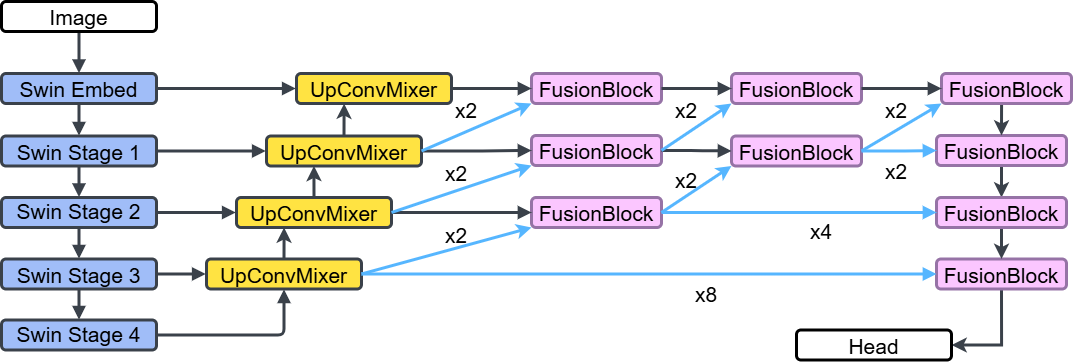}
    \caption{\textbf{Figure: Overall Architecture of GLOD.} The network consists of a Swin Transformer encoder (blue blocks), a custom UpConvMixer decoder (yellow blocks, see \Cref{ssec:upconvmixer}), and Fusion blocks (pink blocks, from HRNet, see \Cref{ssec:fusion_block}) for multi-resolution feature fusion. Blue arrows denote bilinear upsampling with scale factors (\(\times2\), \(\times4\), \(\times8\)). The final detection is performed by the prediction head. The architecture is optimized for detecting small objects in large images.}
    \label{fig:model_architecture}
\end{figure*}

The detection of objects in high-resolution satellite imagery presents significant computational and algorithmic challenges. Traditional approaches often rely on image cropping or multi-pass models to handle the vast data volumes, which are computationally intensive and memory-demanding. These methods struggle to capture global context and long-range dependencies, particularly in detecting tiny or densely packed objects.

Recent advances in transformer-based models have shown promise in capturing global relationships and modelling dependencies over long distances \cite{Caron_2021_ICCV,DBLP:conf/iclr/DosovitskiyB0WZ21, He_2022_CVPR, pmlr-v139-touvron21a}. However, their scalability to high-resolution images remains a significant challenge due to computational intensity. Convolutional Neural Networks (CNNs), while efficient for local feature extraction \cite{7410526, girshick14CVPR, 10.1007/978-3-319-46448-0_2, redmon2016you, NIPS2015_14bfa6bb, zhou2019objects}, are limited in their ability to capture global context.

\paragraph{Hypothesis}
We hypothesize that transformers, known for their ability to model complex relationships \cite{vaswani2017attention}, are better suited for feature extraction in satellite imagery. Specifically, we posit that transformers can more effectively model the dependencies between objects compared to traditional CNN-based approaches. This hypothesis is based on empirical observations in satellite imagery where objects exhibit strong spatial priors (e.g., buildings are adjacent to roads in xView). Transformers, with their self-attention mechanisms, are suited to exploit such priors.

To validate our hypothesis, we propose Global-Local Object Detector (GLOD), a transformer-first architecture that replaces CNN backbones with a Swin Transformer \cite{Liu_2021_ICCV, Liu_2022_CVPR} for end-to-end feature extraction (\Cref{fig:model_architecture}). This design is tailored for datasets like xView \cite{lam2018xview}, where global context and spatial relationships between objects are critical. Our UpConvMixer block addresses the upsampling challenge for tiny objects by combining asymmetric fusion and CBAM attention \cite{10.1007/978-3-030-01234-2_1} to preserve spatial details, while the Fusion Block merges multi-scale features to handle objects of varying sizes (\Cref{fig:model_architecture}).


Our main contributions are as follows:
\begin{itemize}
    \item We introduce GLOD, a novel architecture that combines the strengths of transformers for global dependency modelling with the efficiency of CNNs for local feature refinement.
    \item We propose the UpConvMixer block, a refinement module that integrates asymmetric fusion, separable convolutions, a CBAM attention module and PixelShuffle operations \cite{Shi_2016_CVPR} for robust feature upsampling.
    \item We introduce the Fusion Block, a module that progressively merges features from lower to higher UpConvMixer blocks, enhancing the model's ability to detect objects of varying sizes.
\end{itemize}

\section{Related Work}
\label{sec:related_work}

\paragraph{Foundation Models in Computer Vision.}
Early object detection methods, such as Faster R-CNN \cite{NIPS2015_14bfa6bb} and YOLO \cite{redmon2016you}, relied on anchor-based networks and hierarchical feature extraction. While effective for lower-resolution images, these methods struggle with high-resolution imagery due to their limited ability to capture global context. The Feature Pyramid Network (FPN) \cite{Lin_2017_CVPR} addressed this by constructing a pyramidal feature hierarchy, enabling multi-scale object detection.

\paragraph{Transformer-Based Methods and Hybrid Architectures.}
To better capture global dependencies, transformer-based models such as DETR \cite{10.1007/978-3-030-58452-8_13} have been introduced, removing the need for handcrafted anchors and enabling end-to-end training. However, despite their conceptual elegance, these models remain difficult to scale to high-resolution images. To address this, hybrid architectures like TransUNet \cite{chen2024transunet} combine transformers with CNNs, leveraging the global context modelling of the former and the spatial efficiency of the latter. These hybrid approaches have directly inspired our work on GLOD, which seeks to enhance object detection in high-resolution imagery.

Building on the DETR paradigm, several specialized variants have been proposed to tackle the limitations in detecting objects. For instance, DQ-DETR \cite{10.1007/978-3-031-73116-7_17} introduces a dynamic query mechanism, while DNTR \cite{Liu2024} enhances multi-scale fusion via a denoising FPN. These approaches underscore the continued need for architectural innovation. Yet, improving the underlying loss functions and evaluation metrics is equally critical to advancing detection performance.

\paragraph{Loss Functions and Evaluation Metrics.}
Loss functions play a crucial role in object detection tasks by guiding the training process. Traditional metrics such as Mean Squared Error (MSE) or L1 distance provide simple geometric approximations but often fail to capture localization accuracy. More sophisticated alternatives like Intersection over Union (IoU) \cite{zheng2020diou, Rezatofighi_2019_CVPR} have been developed to address this. For small and tiny objects, recent innovations like Dot Distance (DotD) \cite{Xu_2021_CVPR} yield better gradient signals, while distribution-based metrics such as Normalized Wasserstein Distance (NWD) \cite{Xu2022} improve robustness in noisy or overlapping regions.

On the classification side, techniques like Focal Loss \cite{Lin_2017_ICCV} mitigate class imbalance by emphasizing hard-to-classify examples. These improvements in loss formulation have significantly boosted model robustness and accuracy, especially in challenging detection scenarios.

\paragraph{Diffusion-Based Methods.}
More recently, diffusion-based methods have emerged as an alternative paradigm for object detection. Rather than relying on fixed queries or anchors, these approaches model detection as a generative denoising process \cite{Chen_2023_ICCV}, iteratively refining bounding boxes over time. By framing object detection as a distributional sampling problem, diffusion models hold promise in managing complex scenes with dense or overlapping objects. As this area of research continues to grow, diffusion-based strategies are likely to become an important component of the next generation of object detectors.

%% file: sec/3_methodology.tex
\section{Methodology}
\label{sec:methodology}

In this section, we present the methodology behind our Global-Local Object Detector (GLOD). We provide an overview of the GLOD architecture, detailing its backbone, convolutional neck, and the innovative UpConvMixer block. We also discuss the integration of a CenterNet-inspired head for precise object localization. This section aims to provide a comprehensive understanding of the design choices and mechanisms that enable GLOD to achieve robust and efficient object detection in high-resolution contexts.

\subsection{Architecture Overview}
\label{ssec:model_architecture}

The proposed model architecture is designed to efficiently leverage both the global context and local details of high-resolution images, with a fixed size of $3072 \times 3072$. The backbone of the model is a Swin Transformer \cite{Liu_2021_ICCV}, which is particularly well-suited for capturing long-range dependencies and hierarchical feature representations. Its ability to partition the input into non-overlapping windows ensures computational efficiency, even for high-resolution inputs. Furthermore, the hierarchical structure of the Swin Transformer enables the extraction of multi-scale features, which are crucial for capturing semantic information at varying levels of detail. This design choice contributes to the model’s high learning capacity, as transformers have been demonstrated to outperform convolutions in learning capability \cite{NEURIPS2021_20568692}.

The model incorporates a convolutional neck consisting of four cascaded upsampling modules to process and refine the hierarchical features extracted by the Swin Transformer. These modules aggregate features from different stages of the transformer, enabling a combination of coarse and fine-grained information. To further enhance the quality of the feature maps, the outputs from earlier layers of the Swin Transformer are concatenated with deeper UpConvMixer blocks, a mechanism inspired by U-Net architectures \cite{Ronneberger2015}. This strategy allows the model to effectively propagate low-level spatial details and high-resolution features into the deeper layers, ensuring that critical information is preserved across all scales.

The head of the model is inspired by CenterNet \cite{zhou2019objects}, a framework that treats object detection as a key-point estimation task. By representing objects as centre points, this approach avoids the complexities of traditional bounding box regression, making it particularly effective for scenarios involving dense or overlapping objects. This complements the rich feature representations provided by the neck, allowing the model to output precise localisation predictions. We also believe that the CenterNet method is one of the best ways of not relying on a maximum number of objects.

\subsection{UpConvMixer}
\label{ssec:upconvmixer}

\begin{figure}
    \centering
    \includegraphics[width=0.5\linewidth]{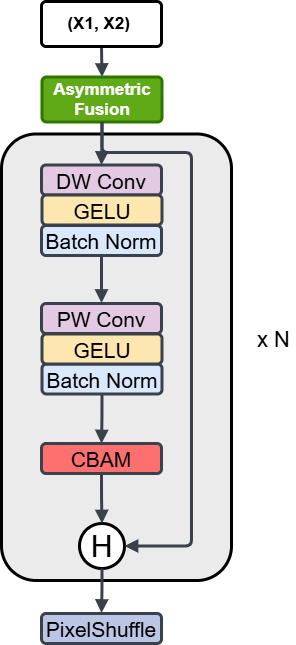}
    \caption{\textbf{Architecture of the UpConvMixer (UCM) block.} The process begins with the concatenation of inputs \(X_1\) and \(X_2\), followed by an asymmetric fusion. This is followed by a series of operations repeated \(N\) times, including depthwise (DW) convolution, GELU activation, batch normalisation, pointwise (PW) convolution, GELU activation, batch normalisation, and a CBAM attention module. The Highway module, denoted by \(H\), is defined as \(g \cdot h + (1 - g) \cdot x\). The final operation is a PixelShuffle, which doubles the spatial resolution. We choose \(N=3\).}
    \label{fig:upconvmixer}
\end{figure}

Within the neck, the UpConvMixer (UCM) blocks (\Cref{fig:upconvmixer}) serve as refinement modules that use depthwise atrous convolutions \cite{Chen2017RethinkingAC} followed by pointwise convolutions to mix the channels, as found in ConvMixer \cite{Trockman2022} or in ResMLP with their patch communication \cite{touvron2021resmlp}.

\begin{figure}
    \centering
    \includegraphics[width=0.65\linewidth]{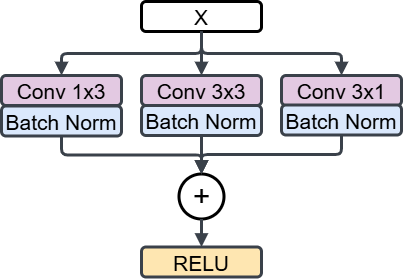}
    \caption{\textbf{Asymmetric Fusion Block.} The input \(X\) is processed through three parallel convolutional paths with different kernel sizes: \(1 \times 3\), \(3 \times 3\), and \(3 \times 1\). Each path is followed by batch normalisation. The outputs of these paths are summed and passed through a ReLU activation function to produce the final fused feature.}
    \label{fig:asymmetric_conv}
\end{figure}

Each block begins with an asymmetric fusion \cite{10530458} (\Cref{fig:asymmetric_conv}, \Cref{eq:asymmetric_fusion}), which integrates the concatenation of \(X_1\) and \(X_2\), denoted as \(X\), through three parallel convolutional paths of different kernel sizes, each followed by batch normalisation (BN) and a ReLU activation. The three core sizes \(1\times3\), \(3\times3\), and \(3\times1\) are used to capture spatial patterns vertically and horizontally, enabling a richer fusion of features.

\begin{equation}
AF_X
= \sigma_{\mathrm{ReLU}}\Bigl(
    \sum_{k\in\{1\times3,\,3\times3,\,3\times1\}}
      \mathrm{BN}\bigl(k\ast X\bigr)
\Bigr)
\label{eq:asymmetric_fusion}
\end{equation}

where \(\sigma_{\mathrm{ReLU}}\) denotes the rectified linear unit. Let \(AF_X\) be the output of \Cref{eq:asymmetric_fusion}. This fused feature is then refined by a series of operations repeated \(N\) times, followed by a CBAM attention module and a Highway module introduced by Srivastava \etal \cite{srivastava2015training} before upsampling:

\begin{equation}
\begin{aligned}
    X_3^{(0)} &= AF_X; \quad \forall i \in [1;N] \\
    X_1^{(i)} &= \sigma_{\mathrm{GELU}}\bigl(\mathrm{BN}(\mathrm{D}^k(X_3^{(i-1)}))\bigr),\\
    X_2^{(i)} &= \sigma_{\mathrm{GELU}}\bigl(\mathrm{BN}(\mathrm{P}(X_1^{(i)}))\bigr),\\
    X_3^{(i)} &= g \cdot X_2^{(i)} + (1 - g) \cdot X_3^{(i-1)}\\
    Y_{UCM}   &= \mathrm{PixelShuffle}(X_3^{(i)}).
\end{aligned}
\label{eq:upconvmixer}
\end{equation}

Here, \(g\) denotes the gating function, basically a pointwise convolution and a sigmoid; \(\mathrm{D}^k\) and \(\mathrm{P}\) denote depthwise and pointwise convolutions respectively with kernel \(k\), known as separable convolutions. We use \(k=3\) for the depthwise convolutions. The PixelShuffle operation efficiently increases spatial resolution without introducing artifacts. The overall method can also be found in MAFF-HRNet \cite{che2023maff}, with the difference that what they call ESCA is actually a CBAM block and that we have placed it at the end of the block in order to filter out the most important features after the transformations.

The number of layers increases with repetition \(N\), and we found during the design that the network could stagnate. We thought this was due to the design of the network itself, becoming too deep and information being lost as it went along \cite{Tishby2015}. We solved this problem by replacing the residual block with a Highway module. The Highway module between \(X_3^{(i-1)}\) and \(X_2^{(i)}\) ensures better gradient flow and information retention. Our experience shows that using the Highway module results in 10\% less loss than using a conventional residual block.

\subsection{Fusion Block}
\label{ssec:fusion_block}

Initially, we placed the network head at the end of the last UCM block. But we soon realised that large objects were not being detected. The most viable hypothesis we investigated was that the information from these objects was in the first UCM blocks, at the output of the Transformer. However, this information is mixed, transformed and passed to the higher blocks without participating directly in the final heatmap. We used a technique borrowed from Wang et al. \cite{wang2020deep}, modified by He \etal \cite{he2022improved} to build our Fusion Block.

The operation is quite simple: the Fusion Block takes as input the outputs of two unified UCMs. The lower-resolution output is first transformed by a pointwise convolution to match the number of channels of the higher-resolution output, then it is upsampled by a factor \(f\) using bilinear interpolation to match the higher resolution. The higher-resolution output is simply passed through a pointwise convolution, retaining the same number of channels. The two results are then added together, and the final result is activated by a GELU function.

\subsection{Loss}
\label{ssec:loss}

\paragraph{Classification.}
To address the inherent class imbalance in object detection tasks, particularly the foreground-background imbalance, we adopt the modified Focal Loss as used in CenterNet. This loss function is defined as:
\begin{equation}
\label{eq:focal-loss}
    L_{cls} = -\frac{1}{N} \sum_{xyc}
\begin{cases}
    (1 - \hat{Y}_{xyc})^\alpha \log(\hat{Y}_{xyc}) \\
    (1 - Y_{xyc})^\beta (\hat{Y}_{xyc})^\alpha \log(1 - \hat{Y}_{xyc}) 
\end{cases}
\end{equation}

Here, $\hat{Y}_{xyc}$ represents the predicted value at position \((x, y)\) for class \(c\), and \(Y_{xyc}\) is the ground truth. 
We use \(\alpha = 2\) and \(\beta = 4\), which are tuned to our specific use case where objects are often very small, sometimes reduced to just one pixel after applying the downsampling factor.

\paragraph{Box‑offset and size regression.}
We optimise two distinct Smooth-L1 objectives — one for the sub‑pixel offsets $(\Delta x,\Delta y)$ and another for the object size $(w,h)$. As a reminder, the Smooth-L1 function is defined as :

\begin{equation}
    \mathcal{L}_{\text{SL1}} = \begin{cases}
        0.5 (x - y)^2 / beta, & \text{if } |x - y| < beta \\
        |x - y| - 0.5 * beta, & \text{otherwise }
        \end{cases}
\end{equation}

\paragraph{Overall objective.}
The network is trained to minimise
\begin{equation}
    \mathcal{L}_{\text{total}} =
    \lambda_{\text{cls}}\mathcal{L}_{\text{cls}}
    + \lambda_{\text{off}}\mathcal{L}_{\text{off}} + \lambda_{\text{size}}\mathcal{L}_{\text{size}}
\end{equation}
with \(\lambda_{\text{cls}} = 1\), \(\lambda_{\text{off}} = 1\) and \(\lambda_{\text{size}}=1\).

%% file: sec/4_experiments.tex
\section{Experiments}
\label{sec:experiments}

We evaluate our approach through a series of experiments on the xView dataset, focusing on architectural design choices and the impact of class imbalance.

\subsection{Dataset}
\label{ssec:dataset}

\paragraph{Summary}
The xView dataset \cite{lam2018xview} is one of the largest and most comprehensive annotated datasets for object detection in satellite imagery. It consists of 847 high-resolution images captured by the WorldView-3 satellite, with a resolution of 0.3 meters per pixel. These images span a wide variety of object classes, including vehicles, buildings, aircraft, and maritime vessels, and contain over one million annotated object instances across 60 different classes.

\begin{figure}
    \centering
    \includegraphics[width=\linewidth]{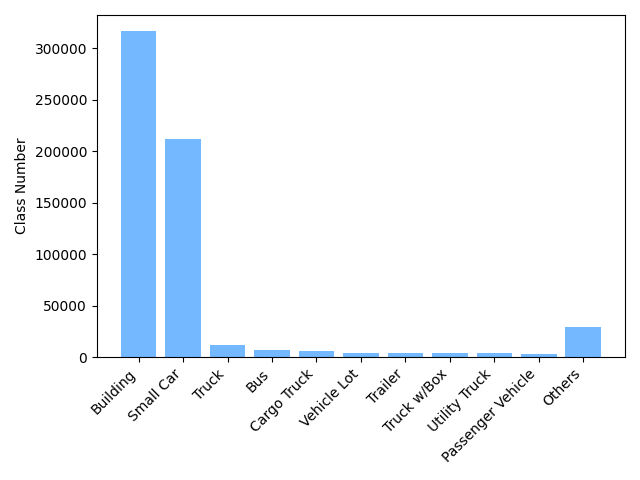}
    \caption{\textbf{Class imbalance in the xView dataset.} The distribution of instances per class is shown, highlighting the significant disparity between frequent classes (e.g., "Small Car" and "Buildings") and rare classes which are grouped together in one category for illustration purposes.}
    \label{fig:class_imbalance}
\end{figure}

\paragraph{Class Imbalance}
The xView dataset exhibits significant class imbalance, with frequent classes like \textit{Small Car} and \textit{Building} dominating, while rare ones such as \textit{Railway Vehicle} have very few instances (\Cref{fig:class_imbalance}). This skews model performance toward common categories and leads to confusion between visually similar classes, especially among vehicle or ship types. 

\subsection{Implementation}
\label{ssec:implementation}

We implement our model using the \textit{Transformers} library \cite{wolf-etal-2020-transformers} and \textit{PyTorch} \cite{paszke2019pytorchimperativestylehighperformance}, leveraging their efficient APIs for transformer-based architectures and deep learning pipelines.

Consistency in image processing is ensured by resizing all images to 3072×3072 pixels, matching the architecture's fixed input size (multiple of the 16-pixel patch size) while preserving sufficient spatial resolution for small object detection. This resolution was determined through empirical analysis to balance memory constraints and detection performance, as it maintains adequate resolution at the Swin Transformer's fourth stage output while remaining computationally feasible.

Consistency in image processing is ensured by resizing all images to 3072×3072 pixels, matching the architecture's fixed input size (a multiple of the 16-pixel patch size) while preserving sufficient spatial resolution for small object detection. This choice of resolution is intentional: resizing the original high-resolution satellite imagery to smaller dimensions such as 800×800 would result in the disappearance or severe degradation of tiny objects like cars, many of which occupy only a few pixels in the native scale. Maintaining a large input size is therefore critical to preserving these fine-grained spatial details.


An extensive data augmentation pipeline enhances the model's robustness and generalisation capabilities, simulating varied lighting conditions, adjusting contrast, and ensuring diversity in spatial orientation. Techniques include greyscale conversion (to reduce lighting effects), solarisation (threshold = 192) for contrast variation, histogram equalisation to balance intensities, and random flips. We intentionally avoid aggressive spatial or geometric augmentations (e.g., elastic deformations, severe blurring, synthetic occlusion) to preserve object shape priors, which are critical in domains like vehicle detection. In satellite imagery, semantic shape consistency is often more important than texture diversity. We also exclude MixUp \cite{zhang2018mixup} and Mosaic \cite{ge2021yoloxexceedingyoloseries} augmentations, which alter spatial object relationships, as our hypothesis is that the Transformer encoder can leverage natural object co-occurrence and spatial context (e.g., cars on a road) to improve detection. Applying these techniques would disrupt the original contextual priors, limiting the model’s ability to exploit such dependencies.

Input features are standardised through normalisation using mean and standard deviation values derived from the ImageNet 1K dataset. This normalisation helps to standardise the data distribution, which is beneficial for training stability and convergence. We deliberately avoid using pre-trained models like Swin Transformers on ImageNet as we think that the feature maps learned from ImageNet are not directly transferable to satellite images. Instead, we opted for training the model from scratch to ensure a robust foundation suited specifically to the characteristics of satellite imagery. We split our dataset to 85/15, as we never had a response from the xView Challenge team to obtain the true test set.

All experiments are conducted on an NVIDIA RTX 4090 GPU, running Ubuntu 24.04.1 LTS with an Intel i9-13900K processor and 64GB of RAM. 

\subsection{Training}
\label{ssec:training}

\paragraph{Training process.} We train the model during 42k steps on xView using AdamW with a learning rate $\alpha$ of $5 \times 10^{-5}$, $\beta_1 =0.9$, $\beta_2=0.999$, and $\epsilon=10^{-8}$, optimising for robustness and class balance. We use Cosine with Warm Restarts to encourage exploration during early training phases, which is especially important when dealing with highly imbalanced data that may trap the model in biased local minima. We choose 10 epochs per cycle. A small per-device batch size of $3$ with gradient checkpointing allows training large-resolution inputs without exceeding GPU memory, while the accumulated real batch size of 24 stabilises gradient estimates for this high-variance detection task.

Key training parameters are selected to optimise performance and address class imbalance. The minimum radius for positive sample assignment is set to 1, allowing for more precise and flexible matching of predicted centres to ground-truth objects. To manage the imbalance between foreground and background samples, a negative sampling ratio of 2\% is applied. This ensures that the model remains focused on the most informative examples, improving overall detection accuracy.

\paragraph{Evaluation Metrics.}
We evaluate model performance using two standard object detection metrics: mean Average Precision at IoU thresholds of 0.5 (mAP50) and 0.75 (mAP75) \cite{10.1007/978-3-319-10602-1_48}. These metrics are widely used in remote sensing and dense object detection tasks, such as xView, to reflect both coarse and fine localization performance.

\textbf{mAP50} measures mean Average Precision using an IoU threshold of 0.5, emphasizing detection coverage and tolerance to moderate localization errors. \textbf{mAP75} applies a stricter 0.75 threshold, rewarding precise bounding boxes—crucial for small or overlapping objects in satellite imagery.

Both mAP50 and mAP75 are defined as:

\begin{equation}
    \text{mAP}_\tau = \frac{1}{N} \sum_{i=1}^{N} \text{AP}_{i,\tau}
\end{equation}

where \( \tau \in \{0.5, 0.75\} \) is the IoU threshold, \( N \) is the total number of object classes, and \( \text{AP}_{i,\tau} \) is the Average Precision for class \( i \) at threshold \( \tau \).

Reporting both metrics enables a comprehensive evaluation of detection coverage and spatial localization accuracy, which is critical in challenging datasets like xView.

\subsection{Results}
\label{ssec:results}

We evaluate GLOD on the xView test set (\Cref{tab:results}), comparing it to CNN and Transformer-based baselines under equivalent training conditions.

GLOD achieves state-of-the-art performance across all reported metrics. Notably, the mAP surpasses previous best results, with particularly significant improvements over both traditional CNN backbones by 4.63 mAP50 points (32.95 vs 28.32) and Transformer-based models by 3.39 mAP50 points (32.95 vs 29.56).

To further characterise the quality of the output heatmaps, beyond classification and localisation, we also evaluate the Peak Signal-to-Noise Ratio (PSNR) between predicted and ground-truth heatmaps, following the approach introduced in DNTR \cite{Liu2024}. PSNR serves as a proxy for heatmap fidelity, quantifying how sharply and accurately the model localises small objects, which often suffer from spatial diffusion in dense detection tasks. High PSNR values indicate less noise and more confident, spatially precise activations.

\begin{figure}
    \centering
    \includegraphics[width=\linewidth]{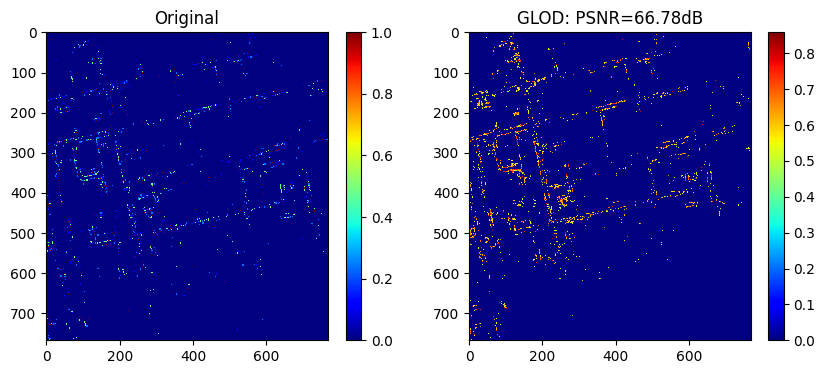}
    \caption{\textbf{Comparison of detection heatmaps between ground truth and GLOD predictions.} Each ground truth object is encoded as a 2D Gaussian peak centred on its location, with a maximum intensity of 1 (left). We report the PSNR (logarithmic scale) between the predicted (right) and ground truth heatmaps as a quantitative measure of spatial fidelity. A higher PSNR reflects more precise and less noisy predictions.}
    \label{fig:heatmaps_psnr}
\end{figure}

DNTR previously reported the best PSNR for car detection (58.0 dB), while GLOD achieves 66.78 dB (\Cref{fig:heatmaps_psnr}), representing a substantial improvement of 9 dB absolute gain over prior results. This indicates that GLOD not only detects more objects but also produces sharper, more precise heatmaps, particularly important for preserving small-object features in remote sensing.

The performance trends indicate that GLOD addresses two critical limitations of previous methods: small object detection under scale variance, which Transformer-based models typically struggle with due to limited inductive biases, and efficient spatial detail preservation during upsampling, where CNN-based models often exhibit blurry activations.

This suggests that GLOD successfully bridges the gap between fine-grained localization and scalability on large-scale datasets such as xView. 

\paragraph{Limitations.}
Despite GLOD's strong performance, several limitations remain. First, the model incurs a higher computational cost compared to purely CNN-based approaches. This overhead stems from the hybrid architecture's reliance on both dense spatial representations and global attention mechanisms, which can hinder deployment in resource-constrained environments or real-time applications.

\begin{figure}
    \centering
    \includegraphics[width=0.8\linewidth]{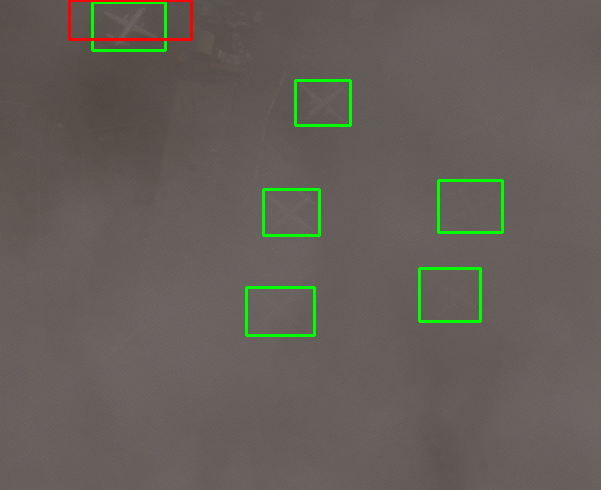}
    \caption{\textbf{Illustration of challenging visibility conditions in aerial imagery.} A human observer might discern the presence of two air-planes behind the clouds. Ground truth in green and predictions in red.}
    \label{fig:clouds}
\end{figure}

Second, the model is sensitive to class imbalance inherent in the xView dataset. Categories with scarce annotations tend to be under-represented in the training signal, which may result in lower recall for rare classes. Additionally, the xView dataset itself poses challenges due to inconsistent and incomplete annotations. In particular, regions with heavy cloud cover — such as those illustrated in \Cref{fig:clouds} — often contain missing or ambiguous labels, which can bias both training and evaluation. Such label noise introduces uncertainty in the supervision signal, limiting the achievable upper bound of detection performance and making it difficult to disentangle model errors from annotation artifacts.

Future work should explore more robust training strategies to mitigate these limitations, such as class-balanced loss functions, uncertainty-aware supervision, and label denoising techniques that account for the imperfect nature of remote sensing datasets.

\begin{figure}
    \centering
    \includegraphics[width=0.8\linewidth]{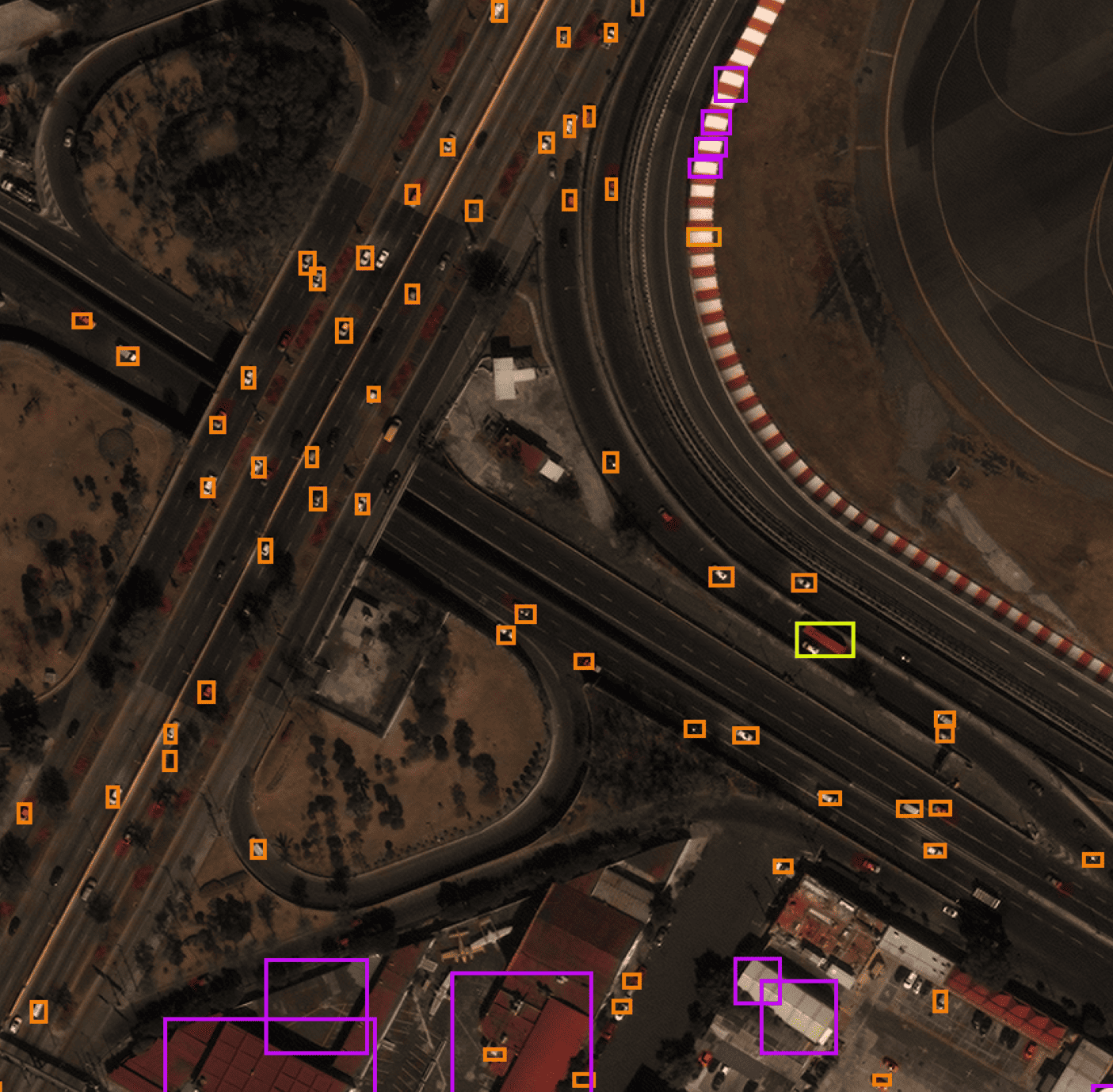}
    \caption{\textbf{Example of detection on the xView test set.} The model correctly detected a large number of vehicles on fast lanes, illustrating good localisation capability in dense environments. However, there are also several false positives for the \textit{building} class (in purple), particularly along the right-hand road, where the model confuses road markings with built structures. These errors generally have a low confidence score, and can therefore be effectively filtered out using a simple post-processing confidence threshold.}
    \label{fig:cars}
\end{figure}

As illustrated in \Cref{fig:prediction_sample} and \Cref{fig:cars}, our model demonstrates strong detection performance. However, we also observe several false positives. These false detections are typically associated with low confidence scores and are visually linked to repetitive patterns in the road markings. This suggests they can be effectively filtered with a conservative confidence threshold, indicating that post-processing can further improve precision without significant recall loss.

\begin{table}
  \centering
  \caption{Comparison of performances on our xView test set. The baseline SSD model, provided by the competition organizers, has its accuracy reflected on the public leader board.}
  \label{tab:results}
    \begin{tabular}{lp{1.5cm}p{1.5cm}}
      \toprule
      \textbf{Method} & \textbf{mAP50 $\uparrow$} & \textbf{mAP75} $\uparrow$ \\
      \midrule
      \multicolumn{3}{c}{CNN-based models} \\
      \midrule
      Baseline (SSD)                                                                & 14.56     & -     \\
      Baseline (SSD-Multires)                                                       & 25.90     & -     \\
      Retina Net                                                                    & 9.70      & -     \\
      FCOS \cite{9897990}                                                           & 17.18     & 12.78 \\
      YOLO11x \cite{yolo11_ultralytics}                                             & 8.42      & 4.28  \\
      DR w/ NWD-RKA \cite{Xu2022, qiao2021detectors}                                & 23.96     & 15.38 \\
      FPN-50 RFL \cite{sergievskiy2019reducedfocalloss1st}                          & 28.32     & -     \\
      \midrule
      \multicolumn{3}{c}{Transformer-based models} \\
      \midrule
      DETR \cite{10.1007/978-3-030-58452-8_13}                                      &  10.64    &  7.89    \\
      DQ-DETR \cite{10.1007/978-3-031-73116-7_17}                                   &  29.56    &  23.94    \\
      DNTR \cite{Liu2024}                                                           &  27.09    &  23.29 \\
      \midrule
      \textbf{GLOD (ours)}                                                          &\textbf{32.95} & \textbf{28.87} \\
      \bottomrule
    \end{tabular}
\end{table}

\subsection{Ablation Study}
\label{ssec:ablation_study}

\paragraph{Influence of the Fusion Block}

To evaluate the impact of the Fusion Block (FB) on detection performance, we conducted an ablation study comparing two variants of our architecture: one without any fusion mechanism (baseline), and one integrating our proposed Fusion Block between intermediate UCM stages.

In the baseline model, the detection head is placed directly at the output of the final UCM block. While this setup yields satisfactory results for small and medium objects, it consistently miss large-scale structures. Visual inspection suggest that crucial information for large objects resides in deeper layers of the backbone, but this information is not sufficiently preserved or propagated to the final detection layer when no fusion is applied.

Integrating the Fusion Block enables the aggregation of multi-scale features by combining spatially precise but semantically shallow features from early layers with semantically rich but spatially coarse features from deeper layers. This fusion facilitates better localization and classification of objects at different scales. The results are recorded in the \Cref{tab:fusion_block_influence}.

\begin{table}
    \centering
    \caption{Ablation study of the Fusion Block (FB) in the GLOD architecture. Detection performance comparison (mAP50 and mAP75) with and without the FB.}
    \label{tab:fusion_block_influence}
    \begin{tabular}{lp{2cm}p{2cm}}
        \toprule
        \textbf{Method} & \textbf{mAP50 $\uparrow$} & \textbf{mAP75} $\uparrow$ \\
        \midrule
        GLOD with FB        & 32.95 & 28.87 \\
        GLOD without FB     & 30.13 & 24.72 \\
        \bottomrule
    \end{tabular}
\end{table}
We therefore retain the Fusion Block as a core component of our architecture, as it consistently improves detection robustness across object scales.

\paragraph{Local Maxima Kernel Size.}
CenterNet eliminates the need for Non-Maximum Suppression (NMS) by using a local maxima filter with a convolutional kernel to identify object centres directly from heatmaps. While the original implementation uses a fixed kernel size of \(3 \times 3\), we conducted an ablation to evaluate how varying this kernel size affects detection performance and object type distribution.

We denote the kernel parameter as \(p\), where the actual filter size is \((2p + 1) \times (2p + 1)\). This effectively controls the spatial extent used to suppress nearby peaks. For all experiments, we apply a top-1000 selection prior to NMS, and report the number and type of retained detections post-filtering.

\begin{itemize}
    \item \textbf{Small kernels (e.g., \(p=0\))} tend to detect many small and densely packed objects such as \textit{Small Car}, but often introduce redundant or overlapping detections.
    \item \textbf{Intermediate kernels (e.g., \(p=1\) to \(p=5\))} balance the detection of both small and medium objects. They yield the highest number of total objects, suggesting a sweet spot for scale-invariant detection.
    \item \textbf{Larger kernels (e.g., \(p \geq 10\))} progressively suppress small object detections in favour of large objects like \textit{Building}.
\end{itemize}

\begin{figure}[h]
    \centering
    \includegraphics[width=\linewidth]{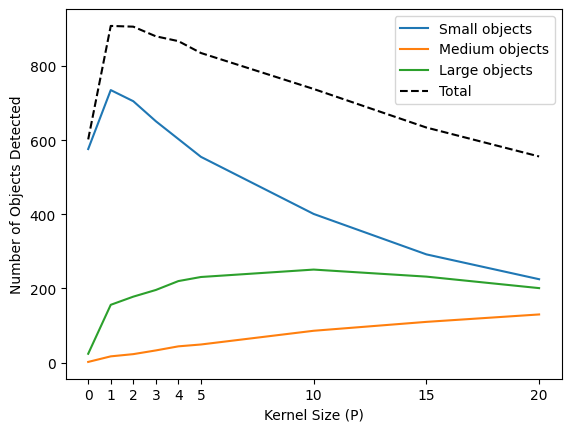}
    \caption{\textbf{Ablation of kernel size \(p\) on detection counts (post-NMS).} Small \(p\) favours dense small-object detection (e.g., cars), while larger \(p\) increasingly biases the output toward larger classes (e.g., buildings).}
    \label{fig:kernel_ablation}
\end{figure}

This trend indicates that the effective receptive field of the NMS kernel acts as an implicit size prior: small kernels favour fine-grained detection, while large kernels coalesce dense regions into larger maxima, benefiting large-object categories (\Cref{fig:kernel_ablation}).

Motivated by these findings, we design a new post-processing pipeline that combines the strengths of different kernel sizes. Specifically, we aggregate detections produced using \(p \in \{0, 1, 10, 20\} \), capturing objects across a wide range of spatial scales from small vehicles to large infrastructure. A standard NMS is then applied to the merged predictions to remove duplicates. This multi-resolution fusion strategy significantly enhances detection diversity and scale robustness without retraining the model.

%% file: sec/5_conclusion.tex
\section{Conclusion}
\label{sec:conclusion}

GLOD demonstrates that transformer-first architectures with asymmetric fusion and multi-scale feature merging can effectively detect objects in high-resolution imagery. Our experiments on xView show 32.95 mAP50, outperforming SOTA by 3.39 points. The architecture's strength lies in preserving spatial details while capturing global context. Future work will explore edge deployment. This approach advances object detection in remote sensing applications.